\documentclass[runningheads]{llncs}
\usepackage{graphicx}
\usepackage[urlcolor=blue citecolor=blue]{hyperref}
\usepackage[misc]{ifsym}

\usepackage{gensymb}
 \usepackage[caption=false]{subfig} 
\usepackage{multirow}
\usepackage{float}

\begin{document}
%

\title{Deep Eyedentification: Biometric Identification using Micro-Movements of the Eye\footnote{This is a pre-print of an article published in Brefeld et al. (Eds.): Machine Learning and Knowledge Discovery in Databases. ECML PKDD 2019, LNCS 11907, pp. 299–314, Springer Nature, 2020, \url{https://doi.org/10.1007/978-3-030-46147-8_18}.}}

\toctitle{Deep Eyedentification: Biometric Identification using Micro-Movements of the Eye}

\titlerunning{Biometric Identification using Eye Movements}

\author{Lena A. J\"ager\inst{1}(\Letter) \and
Silvia~Makowski\inst{1} \and
Paul~Prasse\inst{1} \and
Sascha~Liehr\inst{2}\and
Maximilian~Seidler\inst{1} \and
Tobias~Scheffer\inst{1}}

\tocauthor{ Lena~A.~J\"ager, Silvia~Makowski, Paul~Prasse, Sascha~Liehr, Maximilian~Seidler and Tobias~Scheffer}

\authorrunning{L. J\"ager {\em et al.}}

\institute{Department of Computer Science, University of Potsdam, Potsdam, Germany\\
\email{\{lena.jaeger, silvia.makowski, prasse, maseidler, tobias.scheffer\}@uni-potsdam.de}\and Independet researcher\\
\email{sascha.liehr@gmail.com}}
\maketitle              
\begin{abstract}
We study involuntary micro-movements of the eye for biometric identification. 
While prior studies extract lower-frequency macro-movements from the output of video-based eye-tracking systems and engineer explicit features of these macro-movements, we develop a deep convolutional architecture that processes the raw eye-tracking signal. Compared to prior work, the network attains a lower error rate by one order of magnitude and is faster by two orders of magnitude: it identifies users accurately within seconds.

\end{abstract}

\section{Introduction}
\label{sec:Introduction}
Human eye movements are driven by a highly complex interplay between voluntary and involuntary processes related to oculomotor control, high-level vision, cognition, and attention. 
Psychologists distinguish three types of macroscopic eye movements. 
Visual input is obtained during  {\em fixations} of around 250~ms. {\em Saccades}  are  fast relocation movements of typically  30 to 80 ms between fixations during which visual uptake is suppressed. 
When tracking a moving target, the eye performs a {\em smooth pursuit} \cite{Holmqvist2011}. 

A large body of psychological evidence shows that these macroscopic eye movements are highly individual. For example, a  large-scale study with over 1,000 participants showed that the individual characteristics of eye movements are highly reliable and, importantly, persist across experimental sessions \cite{Bargary2017}.  
Motivated by these findings, macro-movements of the eye have been studied for biometric identification \cite{Bednarik2005,KasprowskiHarkezlak2014}.  
Since macroscopic eye movements occur at a low frequency, long sequences must be observed before movement patterns give away the viewer's identity---a recent study finds that users can be identified reliably after reading around 10 lines of text \cite{Makowski2018}. For use cases such as access control, this process is too slow by one to two orders of magnitude. 

During fixations, the eye additionally performs involuntary micro-movements which prevent the gradual fading of the image that would otherwise occur as the neurons become desensitized to a constant light stimulus \cite{Riggs1952,Ditchburn1952}. 
{\em Microsaccades}  have a duration ranging from 6 to 30~ms 
\cite{Martinez-Conde2004,Martinez-Conde2006,Martinez-Conde2009}. 
Between microsaccades, a very slow {\em drift} 
away from the center of the fixation occurs, 
which is superimposed by a low-amplitude, high-frequency {\em tremor} of approximately 40-100 Hz \cite{Martinez-Conde2004}. 
There is evidence that microsaccades  exhibit measurable individual differences \cite{Poynter2013}, but it is still unclear to what extent drift and tremor vary between individuals~\cite{Ko2016}.

Video-based eye-tracking systems measure gaze angles at a rate of up to 2,000 Hz. Since the amplitudes of the smallest micro-movements are in the order of the precision of widely-used systems, the micro-movement information in the output signal is superimposed by a considerable level of noise. 
It is established practice in psychological research to smooth the raw eye-tracking signal, and to extract the specific types of movements under investigation.
Criteria that are applied for the distinction of specific micro-movements are to some degree arbitrary \cite{Otero2008,Nystrom2016}, and their detection is less reliable~\cite{Ko2016}. 
Without exception, prior work on biometric identification only extracts macro-movements from the eye-tracking signal and defines explicit features such as distributional features of fixation durations and saccade amplitudes.

The additional information in the high-frequency and lower-amplitude  micro-movements motivates us to explore the raw eye-tracking signal for a potentially much faster biometric identification. 
To this end, we develop a deep convolutional neural network architecture that is able to process this  signal. One key challenge lies in the vastly different scales of velocities of micro- and macro-movements. 

The remainder of this paper is structured as follows. Section~\ref{sec-2} reviews prior work. Section~\ref{sec:ProblemSetting} lays out the problem setting and Section~\ref{sec:Architecture} develops a neural-network architecture for biometric identification based on a combination of micro- and macro-movements of the eye. Section~\ref{sec:Experiments} presents experimental results. Section~\ref{sec:discussion} discusses micro-movement-based identification in the context of other biometric technologies; Section~\ref{sec:Conclusion} concludes.

\section{Related Work}\label{sec-2}
There is a substantial body of research on biometric identification using macro-movements of the eye. Most work uses the same stimulus for training and testing---such as a static cross \cite{Bednarik2005},  a jumping point \cite{Kasprowski2004,Kasprowski2005,Rigas2012a,Cuong2012,Srivastava2015}, a face  image \cite{Rigas2012,Galdi2013,Cantoni2015}, or various other kinds of images \cite{Darwish2013}. 
Using the same known stimulus for training and testing opens the methods to replay attacks.

Kinnunen and colleagues present the first approach that uses different stimuli for training and testing and does not involve a secondary task; they identify subjects who watch a movie sequence \cite{Kinnunen2010}. Later approaches use eye movements on novel text to identify readers \cite{Holland2011,Landwehr2014}.

A number of methods have been benchmarked in challenges~\cite{KasprowskiKomogortsevKarpov2012,KasprowskiHarkezlak2014}. All participants in these challenges and all follow-up work~\cite{Rigas2016} present methods that extract saccades and fixations, and define a variety of features on these macro-movements, including distributions of fixation durations and of amplitudes, velocities, and directions of saccades. 
Landwehr and colleagues define a generative graphical model of saccades and fixations \cite{Landwehr2014} from which Makowski and colleagues  derive a Fisher Kernel \cite{Makowski2018}; Abdelwahab {\em et al.}  develop a semi-parametric discriminative model \cite{Abdelwahab2016}. 
All known methods are designed to operate on an eye-gaze sequence of considerable length; for example, one minute of watching a video or reading about one page of text.

\section{Problem Setting}
\label{sec:ProblemSetting}

We study three variations of the problem of biometric identification based on a sequence $\langle (x_0, y_0),\dots, (x_n,y_n)\rangle$ of yaw gaze angles $x_i$ and pitch gaze angles $y_i$ measured by an eye tracker. 
For comparison with prior work, we adopt a {\em multi-class classification} setting. For each user from a fixed population of users, one or more enrollment eye-gaze sequences are available that are recorded while the user is reading text documents. A multi-class classification model trained on these enrollment sequences recognizes users from this population at application time while the users are reading different text documents. Classification accuracy serves as performance metric in this setting.

Multi-class classification is a slight abstraction of the realistic use case in two regards. First, this setting disregards the possibility of encountering a user from outside the training population of users. Secondly, the learning algorithm has to train the model on enrollment sequences of all users. This training would have to be carried out on an end device or a cloud backend whenever a new user is enrolled; this is unfavorable from a product perspective. 

In the more realistic settings of {\em identification} and {\em verification}, an embedding is trained offline on eye-gaze sequences for training stimuli of a population of training identities. At application time, the model encounters users from a different population who may view different stimuli. Users are enrolled by simply storing the embedding of their enrollment sequences. The model identifies a user when a similarity metric between an observed sequence and one of the enrollment sequences exceeds a recognition threshold. 

In the {\em identification} setting, multiple users can be enrolled. Since the ratio of enrolled users to impostors encountered by the system at application time is not known, the system performance has to be characterized by two ROC curves. One curve characterizes the behavior for enrolled users; here, false positives are enrolled users who are mistaken for different enrolled users. The second curve characterizes the behavior for impostors; false positives are impostors who are mistaken for one of the enrolled users. 

In the {\em verification} setting, the model verifies a user's presumed identity. This setting is a special case of identification in which a single user is enrolled. As no confusion of enrolled users is possible, a single ROC curve characterizes the system performance.

\section{Network Architecture}
\label{sec:Architecture}

We transform each eye-gaze sequence $\langle (x_0, y_0),\dots, (x_n,y_n)\rangle$ of absolute angles into a sequence $\langle (\delta^x_1, \delta^y_1),\dots, (\delta^x_n,\delta^y_n)\rangle$ of angular gaze velocities in \degree/s with $\delta_i^x=r(x_i-x_{i-1})$ and $\delta^y_i=r(y_i-y_{i-1})$, where $r$ is the sampling rate of the eye tracker in Hz.

The angular velocity of eye movements differs greatly between the different types of movement. While drift occurs at an average speed of around 0.1-0.4\degree/s and tremor at up to 0.3\degree/s, microsaccades move at a rapid 15 to 120\degree/s and saccades even at up to 500\degree/s \cite{Martinez-Conde2004,Martinez-Conde2009,Otero2008,Holmqvist2011}; there is, however, no general agreement about the exact cut-off values between movement types. 
Global normalization of the velocities squashes the velocities of drift and tremor to near-zero and models trained on such data resort to extracting patterns only from macro-movements. For this reason, our key design element of the architecture consists of independent subnets for {\em slow} and {\em fast} movements which observe the same input sequences but with different scaling.

\begin{figure*}
	\centering
	\includegraphics[trim=.1cm 6.5cm 8.0cm 0.8cm, clip, width=\textwidth]{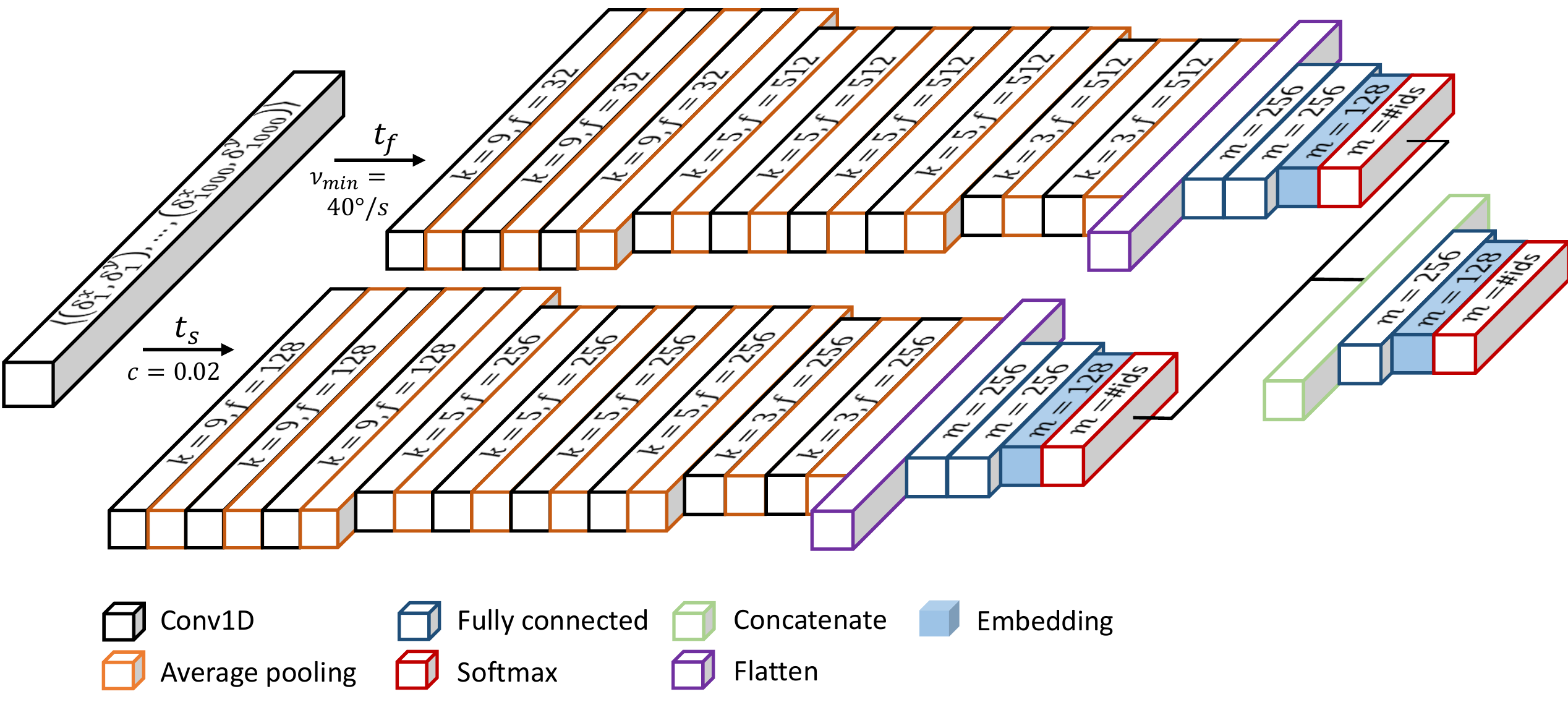}
	\caption{Network architecture. Parameter $c$ denotes the scaling factor of Equation~\ref{eq:trans1}, $\nu_{min}$  the velocity threshold of Equation~\ref{eq:trans2},  \textit{k} the kernel size, \textit{f} the number of filters and \textit{m} the number of fully connected units. Batch normalization and ReLU activation are applied to the output of all convolutional and fully connected layers. All convolutional layers have a stride of 1; all pooling layers have a pooling size of 2 and a stride of 1.}
	\label{fig:architecture}
\end{figure*}

Both subnets have the same number and type of layers; Figure~\ref{fig:architecture} shows the architecture. 
Both subnets process the same sliding window of 1,000 velocity pairs which corresponds to one second of input data, but the input is scaled differently. 
Equation~\ref{eq:trans1} transforms the input such that the low velocities that occur during tremor and drift roughly populate the value range between $-0.5$ and $+0.5$ while velocities of microsaccades and saccades are squashed to values between $-0.5$ and $-1$ or $+0.5$ and $+1$, depending on their direction. The parameter $c$ has been tuned within the range of psychologically plausible values from 0.01 to 0.06. 
\begin{equation}
 t_s(\delta^x_i,\delta^y_i)=(\tanh(c\delta^x_i), \tanh(c\delta^y_i))
    \label{eq:trans1}
 \end{equation}
 
Equation~\ref{eq:trans2}, in which $z(\cdot)$ is the $z$-score normalization, truncates absolute velocities that are below the  minimal velocity $\nu_{min}$
of microsaccades and saccades. Based on the psychological literature, the  threshold $\nu_{min}$ was tuned within the range of 10 to 60\degree/s. 
 \begin{equation}
 t_f(\delta^x_i,\delta^y_i)=\left\{ 
 \begin{array}{l}
 z(0) \ \qquad  \textrm{if } \sqrt{{\delta^x_i}^2+{\delta^y_i}^2} < \nu_{min}\\
(z(\delta^x_i),z(\delta^y_i))   \qquad \textrm{otherwise }
 \end{array}\right.
  \label{eq:trans2}
 \end{equation}

Each subnet consists of 9 pairs of one-dimensional convolutional and average-pooling layers. The model performs a batch normalization on the output of each convolutional layer before applying a ReLU activation function and performing average pooling. Subsequently, the data feeds into two fully connected layers with batch-normalization and ReLU activation with a fixed number of $2^8$ and $2^7$ units, followed by a fully connected layer of $2^7$ units with ReLU activation that serves as embedding layer for identification and verification. For classification and for the purpose of training the network in the identification and verification setting, this is followed by a softmax output layer with a number of units equal to the number of training identities that is discarded after training in the identification and verification settings. 

Figure~\ref{fig:architecture} shows the overall architecture which we refer to as the {\em DeepEyedentification} network.
The output of the subnets is concatenated and flows through a fully connected layer of $2^8$ units and a fully connected layer with $2^7$ units that serves as embedding layer for identification and verification, both with batch normalization and ReLU activation. 
The overall architecture is trained in three steps. The fast and the slow subnets are pre-trained independently and their weights are frozen in the final step where the joint architecture is trained. 

In the identification and verification settings, the final embedding consists of the concatenation of the joint embedding and the embeddings generated by the fast and slow subnets. In this case, the cosine similarity serves as metric for the comparison of enrollment and input sequences.

\begin{table}[t!]
	\centering
	\caption{Parameter space used for grid search: kernel size $k$ and number of filters $f$ of the convolutional layers, the scaling parameter $c$ of  Equation~\ref{eq:trans1} and the velocity threshold $\nu_{min}$ of Equation~\ref{eq:trans2}.} \label{tab:hyperparams_space}
	\begin{tabular}{l|l}
		Parameter     & Search space                     \\ \hline
		$c$           & $\{0.01, 0.02, 0.04, 0.06\}$    \\ \hline
		$\nu_{min}$           & \{$10$\degree/s, $20$\degree/s, $30$\degree/s, $40$\degree/s, $60$\degree/s\}    \\ \hline
		$k$  & $\{3, 5, 7, 9\}$            \\ \hline
		$f$ & $\{32, 64, 128, 256, 512\}$ \\ \hline
	\end{tabular}
\end{table}

\begin{table}[t!]
	\centering
	\caption{Best hyperparameter configuration found via  grid search in the search space shown in Table~\ref{tab:hyperparams_space}.} \label{tab:hyperparams_results}
	\begin{tabular}{l|l|l|l}
		Parameter & Layer                                                                  & Slow subnet                                             & Fast subnet                                            \\ \hline
		$c$       & $t_s$                                                                  & 0.02                                                     & -                                                      \\ \hline
		$\nu_{min}$       & $t_f$                                                                  & -                                                       & 40\degree/s                                                    \\ \hline
		$k$      & \begin{tabular}[c]{@{}l@{}}conv 1-3\\ conv 4-7\\ conv 8-9\end{tabular} & \begin{tabular}[c]{@{}l@{}}9\\ 5\\ 3\end{tabular}       & \begin{tabular}[c]{@{}l@{}}9\\ 5\\ 3\end{tabular}      \\ \hline
		$f$     & \begin{tabular}[c]{@{}l@{}}conv 1-3\\ conv 4-7\\ conv 8-9\end{tabular} & \begin{tabular}[c]{@{}l@{}}128\\ 256\\ 256\end{tabular} & \begin{tabular}[c]{@{}l@{}}32\\ 512\\ 512\end{tabular} \\ \hline
	\end{tabular}
\end{table}

	\section{Experiments}	\label{sec:Experiments}
This section reports on experiments in the settings of multi-class classification, identification, and verification.
All code is available at \url{https://osf.io/ps9qj/}.

\subsection{Data Collection}
We use two data collections for our experiments.
Makowski {\em et al.}~\cite{Makowski2018} have collected the largest eye-tracking data set for which the raw output signal is available. It consists of monocular eye-tracking data sampled at 1,000~Hz from 75 participants who are reading 12 scientific texts of approximately 160 words.  
In order to extract absolute gaze angles, the eye tracker has to be calibrated for each participant. 	
Makowski {\em et al.} exclude data from 13 participants whose data is poorly calibrated. 
Since DeepEyedentification only processes velocities, we do not exclude any data. We refer to this data set as {\em Potsdam Textbook Corpus}.  

The Potsdam Textbook Corpus was acquired in a single session per user.  To explore whether individuals can be recognized across sessions, we collect  an additional data set from 10 participants à four sessions with a temporal lag of at least one week between any two sessions.  
We record participants' gaze using a binocular Eyelink Portable Duo eye tracker at a  sampling rate of 1,000 Hz. 
During each session, participants are presented with 144 trials in which a black point consecutively appears at 5 random positions
on a light gray background 
 on a 38$\times$30~cm monitor (1280$\times$1024~px). The interval in which the point changes  its location varies between trials (250, 500, 1000 or 1500~ms). We refer to these data as {\em JuDo} (Jumping Dots) data set.  We use the Potsdam Textbook Corpus for hyperparameter optimization, and evaluation of the DeepEyedentification network in a multi-class classification and an identification and verification setting, while  we use the much smaller JuDo data set to assess the model's session bias.

\subsection{Reference Methods}

Existing methods for biometric identification using eye movements only operate on macroscopic eye movements; they first preprocess the data into sequences of saccades and fixations and use different features computed from these macro-movements such as fixation duration or saccade amplitude. 
Existing methods that allow different stimuli for training and testing can be classified into i) approaches which  aggregate the extracted  
features over the relevant recording window,  ii) statistical approaches that compute the similarity of scanpaths by applying statistical tests to the distributions of the extracted  features, and iii) 
graphical models that generate sequences of fixation durations and saccade amplitudes.
 As representative aggregational reference method, we choose the model by Holland and Komogortsev (2011) that is specifically designed for eye movements in reading \cite{Holland2011}. 
As statistical reference approaches we use the first model of this kind by Holland and Komogortsev (2013) \cite{Holland2013} and the current state-of-the-art model by  Rigas  {\em et al.}  (2016)~\cite{Rigas2016}. As representative graphical models, we also use the first model of this kind by Landwehr {\em et al.} (2014)~\cite{Landwehr2014} and the state-of-the-art model by  Makowski {\em et al.} (2018)~\cite{Makowski2018}.

\subsection{Hyperparameter Tuning}
\label{sec:hyperparams}
	We optimize the hyperparameters via grid search on one hold-out validation text from the Potsdam Textbook Corpus which we subsequently exclude from the training and testing of the final network; Table~\ref{tab:hyperparams_space} gives an overview of the space of values and Table~\ref{tab:hyperparams_results} the selected values that we keep fixed for all subsequent experiments. 
We vary the kernel sizes and numbers of filters of each subnet independently, but constrain them to be identical within convolutional layers 1-3, 4-7, and  8-9. Moreover, we constrain the kernel size to be smaller or equal and the number of filters to be greater or equal compared to the preceding block.

\subsection{Hardware and Framework}
\label{sec:framework}
		We train the networks on a server with a 40-core Intel(R) Xeon(R) CPU E5-2640 processor and 128 GB of memory and a GeForce GTX TITAN X GPU using the NVidia CUDA platform with Tensorflow version 1.12.0 \cite{tensorflow} and Keras version 2.2.4 \cite{keras}.  
As optimizer, we use Adam \cite{Adam,Reddi2018} with a learning rate of 0.001 for the training of the subnets and 0.0001 for the common layers. All models and submodels are trained with a batch size of 64 sequences.

\subsection{Multi-Class Classification}
This section focuses on the multi-class classification setting in which the model is trained on the Potsdam Textbook Corpus to identify users from a fixed population of 75 users who are represented in the training data, based on an eye-gaze sequence for an unseen text. 

In this setting, data are split across texts, to ensure that the same stimulus does not occur in training and test data. 
We perform leave-one-text-out cross validation over 11 text documents. We study the accuracy as a function of the duration of the eye-gaze signal. For each duration, we let the model process a sliding window of 1,000 time steps and average the scores over all window positions.

The reference models are evaluated on the same splits. They receive preprocessed data as input: sequences are split into saccades and fixations, and  the relevant fixation and saccade features are computed. At test time, these models receive only as many macro-movements as input as fit into the duration. 

\begin{figure}[h!]	
\centering
\includegraphics[width=\textwidth]{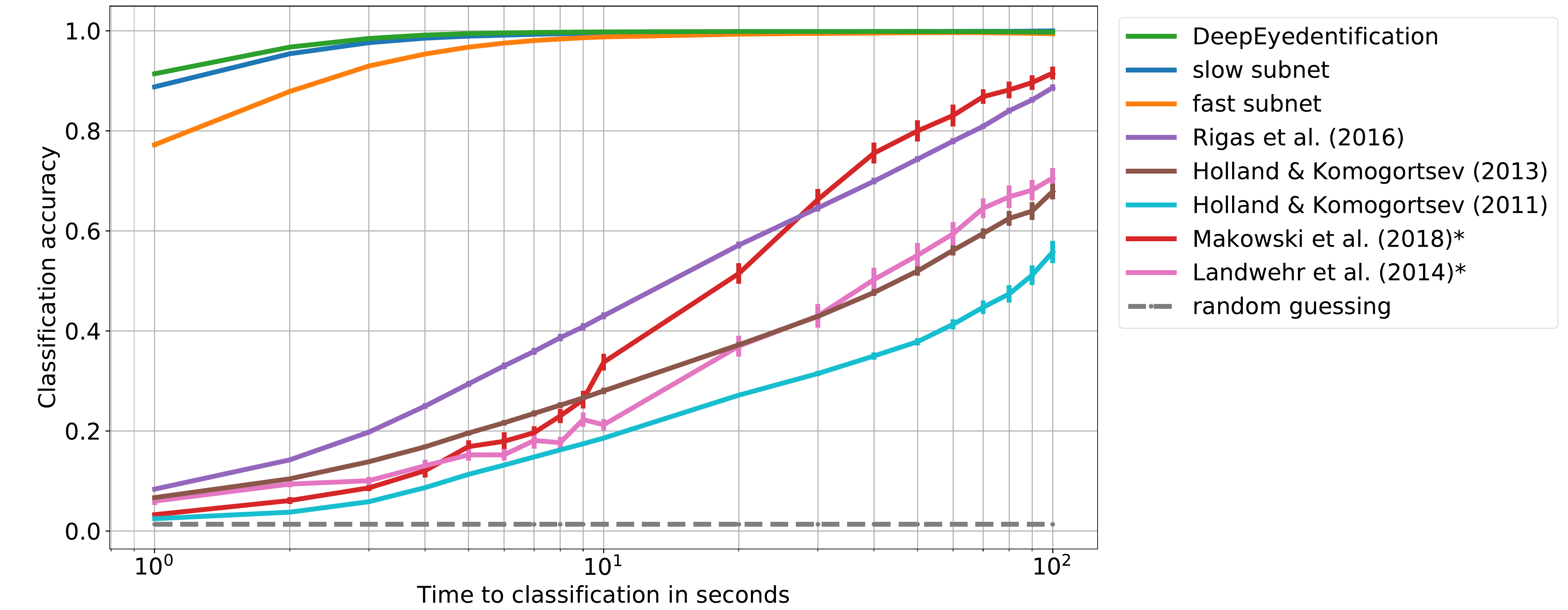}	
	\caption{Multi-class classification on the Potsdam Textbook Corpus.  Categorical accuracy as a function of the amount of available test data in seconds; error bars show the standard error.  
	The models marked with $^{*}$ are evaluated on a subset of the data containing  62 well-calibrated users, all other methods are evaluated on the full data set of 75 readers.} 
	\label{fig:multiclass_classification}
\end{figure}

Figure~\ref{fig:multiclass_classification} shows that for any duration of an input sequence, the error rate of DeepEyedentification is roughly one order of  magnitude below the error rate of the reference methods. 
DeepEyedentification exceeds an accuracy of 91.4\% after one second, 99.77\% after 10 seconds and reaches 99.86\% accuracy after 40 seconds of input, whereas Rigas {\em et al.}~\cite{Rigas2016} reach 8.37\% accuracy after one second and 43.02\% after 10 seconds, and the method of Makowski  {\em et al.}~\cite{Makowski2018} reaches 91.53\% accuracy after 100 seconds of input. We can conclude that micro-movements convey substantially more information than lower-frequency macro-movements of the eye.

The figure also shows that the overall network is significantly more accurate than either of its subnets. The {\em fast subnet}, for which only velocities of microsaccades and saccades are visible while tremor and drift are truncated to zero, reaches an accuracy of approximately 77\%  after one second. The {\em slow subnet}, which perceives the velocities of tremor and drift on an almost-linear scale while the velocities of microsaccades and saccades are squashed by the sigmoidal transformation, achieves roughly 88\% of accuracy after one second.

\subsection{Identification and Verification}

In these settings, the input window slides over the test sequence and an enrolled user is identified (true positive) if and when the cosine similarity between the input window and any window in his enrollment sequence exceeds the recognition threshold; otherwise, the user counts as a false negative. 
A false positive arises when the similarity between a test sequence from an enrolled user (confusion setting) or an impostor (impostor setting) and the enrollment sequence of a different user exceeds the threshold;  otherwise a true negative arises. 
We perform 50 iterations of random resampling on the Potsdam Textbook Corpus. In each iteration, we randomly draw 50 training users and train the DeepEyedentification model on 9 training documents for these users. One text serves as enrollment sequence and one text remains as observation. In the identification setting, a randomly drawn set of 20 of the 25 users who are not used for training are enrolled, and the remaining 5 users act as impostors. In the verification setting, one user is enrolled and 24 impostors remain.

\begin{figure}[htbp!]
	\subfloat[Confusions between 20 enrolled users. \label{fig:conf_20}]{	
		\includegraphics[width=0.48\textwidth, trim={0.7cm 0 2.2cm 2.4cm}, clip]{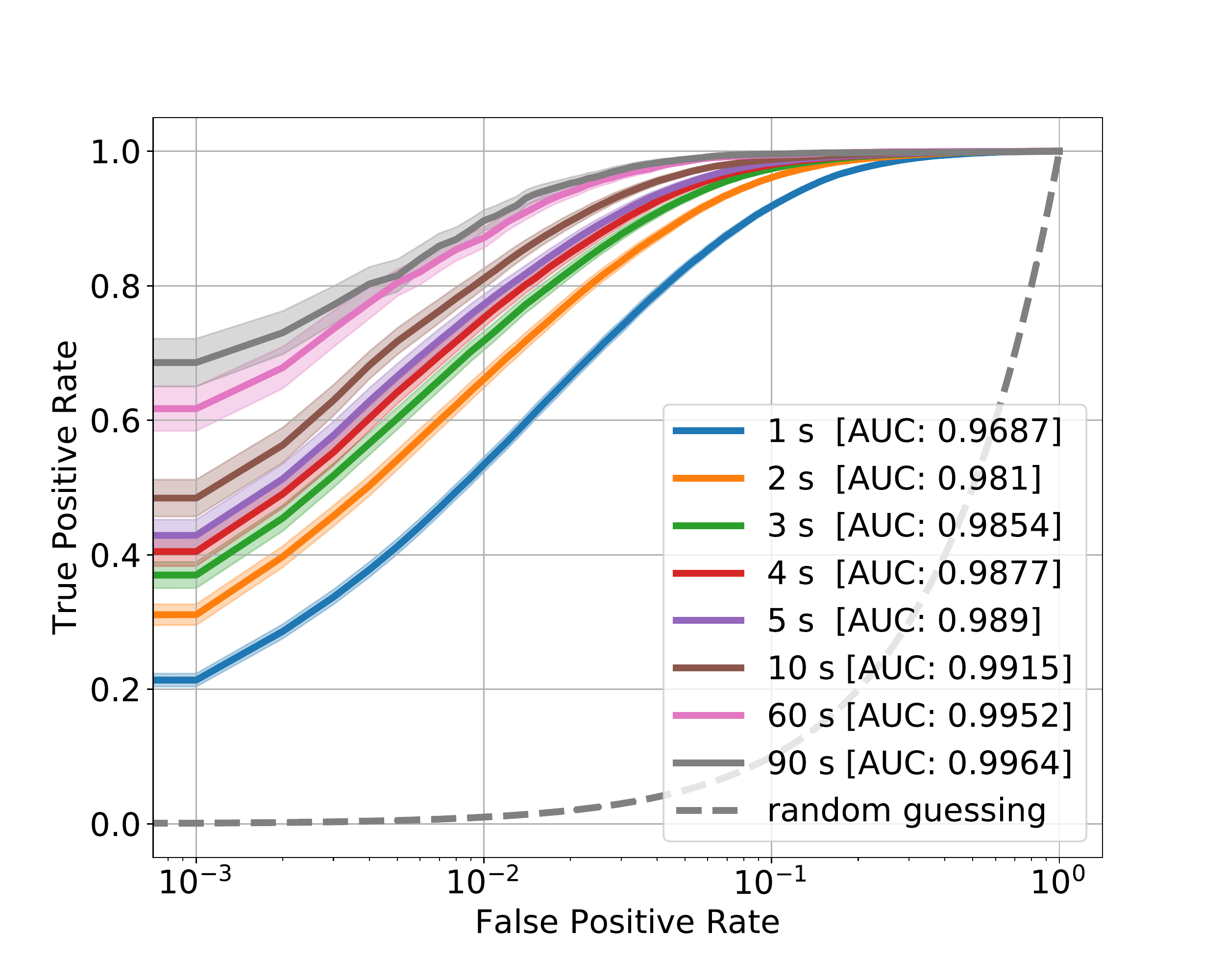}
		}	
		\hfill
	\subfloat[Confusions between an unknown number of impostors and 20 enrolled users.	\label{fig:imp_20}]
	{\includegraphics[width=0.48\textwidth, trim={0.7cm 0 2.2cm 2.4cm}, clip]{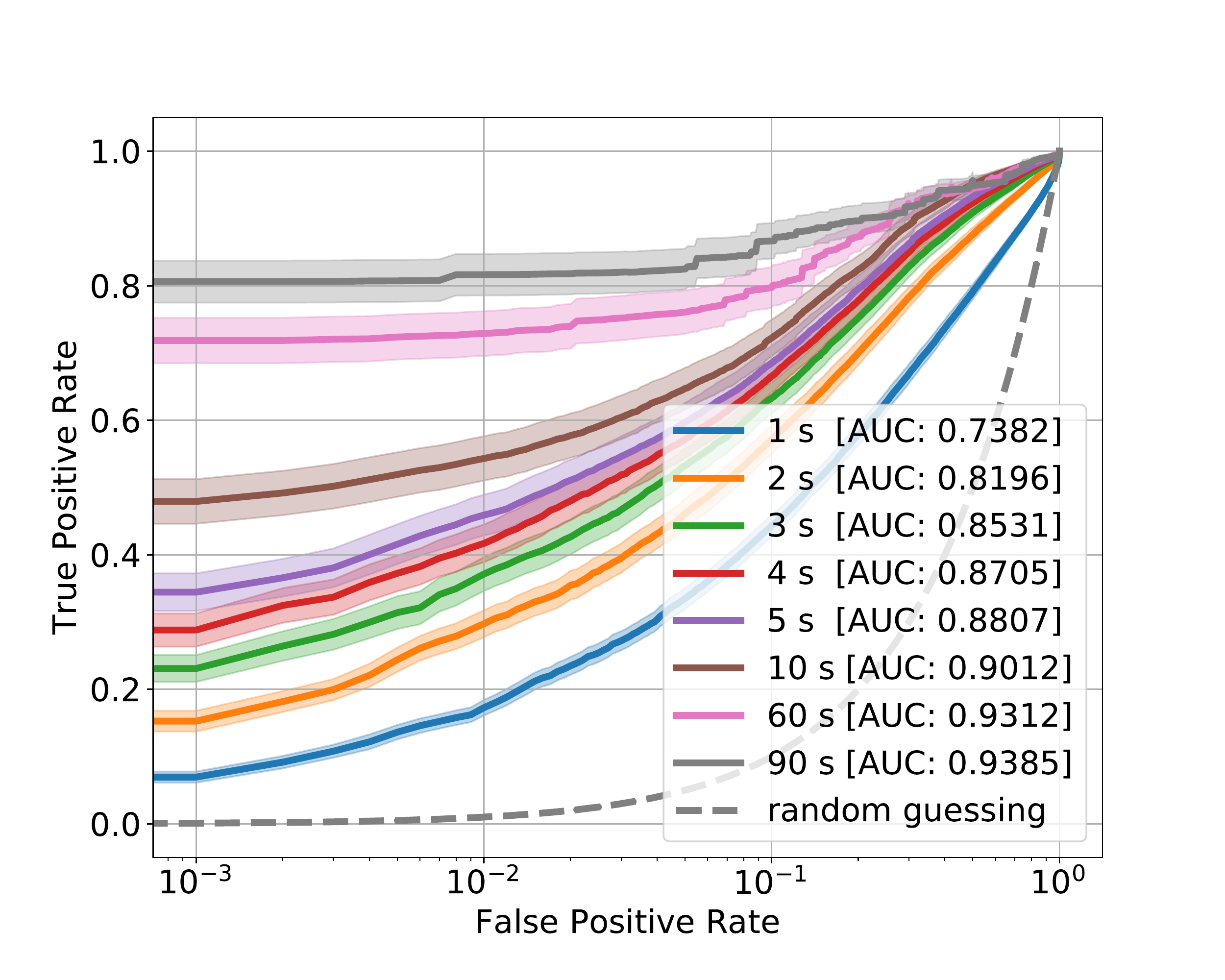}	
	}	
	\caption{Identification on the Potsdam Textbook Corpus. ROC curves for the confusion setting (a) and the impostor setting (b) as a function of the duration of the input signal at application time, both with 20 enrolled users. Error bars show the standard error.}
\end{figure}

\begin{figure}[htbp!]
\centering
\includegraphics[width=0.48\textwidth, trim={0.7cm 0 2.2cm 2.4cm}, clip]{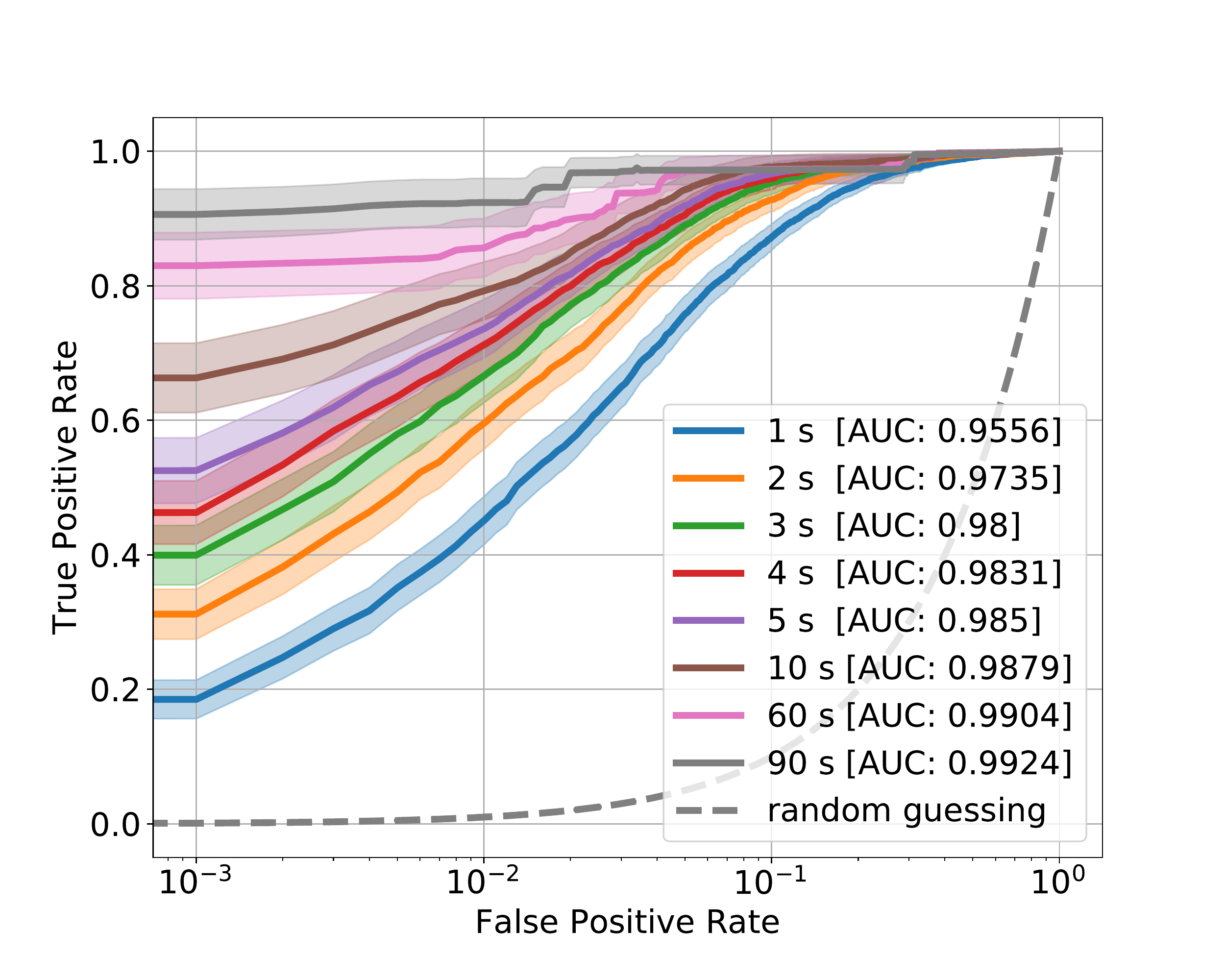}
\caption{Verification on the Potsdam Textbook Corpus. ROC curves for the confusions between one enrolled user and an unknown number of impostors as a function of the duration of the input signal at application time. Error bars show the standard error.}
\label{fig:imp_1}
\end{figure}

For the identification setting, Figure~\ref{fig:conf_20} shows the ROC curves for confusions between the 20 enrolled users on a logarithmic scale. The area under the ROC curve increases from 0.9687 for one second of data to 0.9915 for 10 and 0.9964 after 90 seconds; the corresponding EER values are 0.09, 0.04, and 0.02. Figure~\ref{fig:imp_20} shows the ROC curves for confusions between an impostor and one of the 20 enrolled users; here, the AUC values lie between 0.7382 and 0.9385, the corresponding EER values between 0.31 and 0.1.

\begin{figure}[h!]
\begin{minipage}[b]{0.48\textwidth}%
\centering	
	\includegraphics[width=1.0\textwidth, trim={0.7cm 0 2.2cm 2.2cm}, clip]{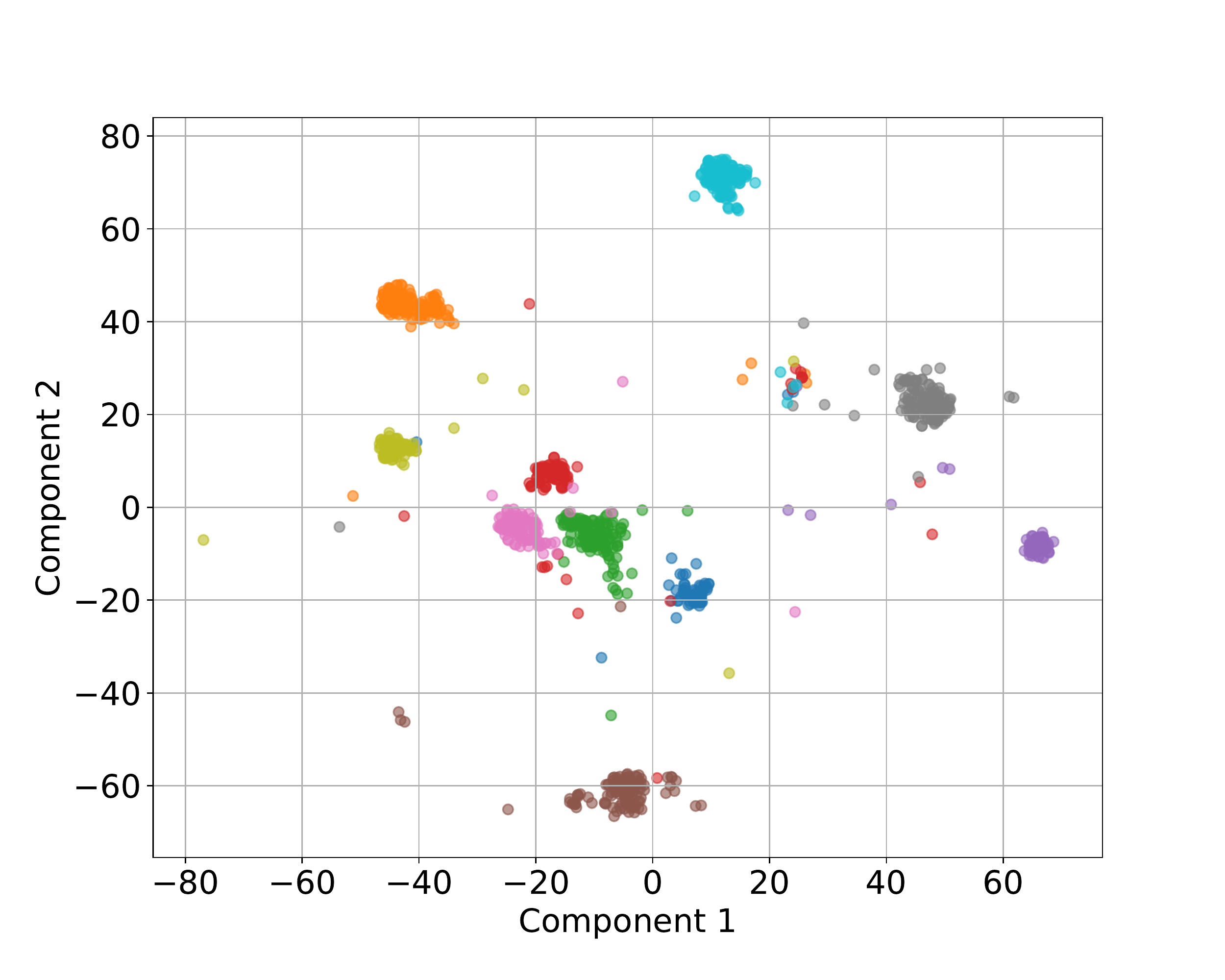}	
	\caption{$t$-sne visualization of the embedding for 10 users.}
	\label{fig:tsne}
\end{minipage}
\hfill
\begin{minipage}[b]{0.48\textwidth}%
\centering	
	\includegraphics[width=1.0\textwidth, trim={0.7cm 0 2.2cm 2.2cm}, clip]{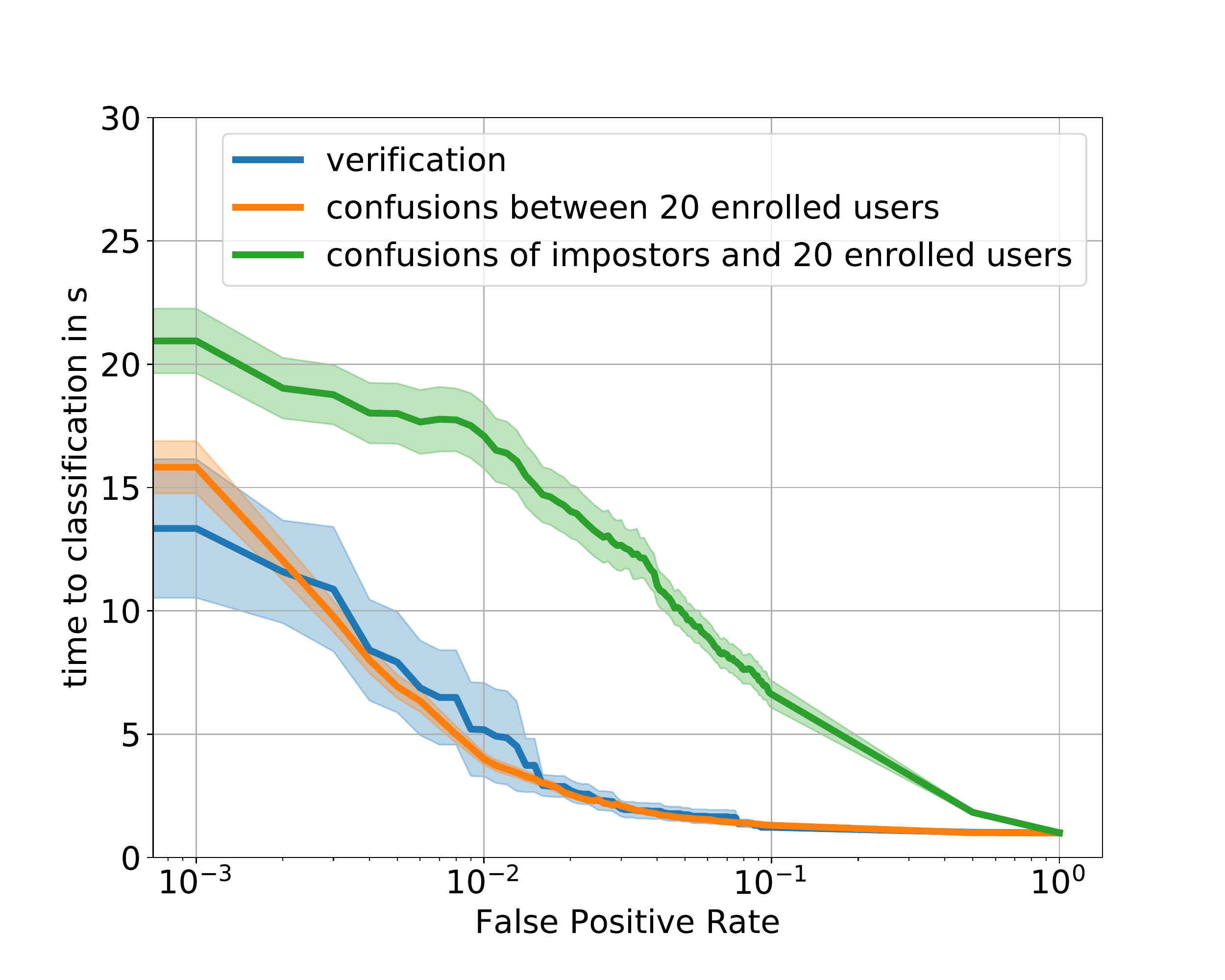}		
	\caption{Time to classification with standard error over false-positive rate.}
	\label{fig:time_to_classification}
\end{minipage}
\end{figure}

Figure~\ref{fig:imp_1} shows the ROC curve for the verification setting. Here, the AUC lies between 0.9556 for one second, 0.9879 for 10, and 0.9924 for 90 seconds. In this setting, each impostor can only be confused with one presumed identity, whereas, in the identification setting, an impostor can be confused with each of the 20 enrolled users.
Figure~\ref{fig:tsne} shows a $t$-SNE visualization~\cite{Maaten2008} that illustrates how the embedding layer clusters 10 users randomly drawn from outside the training identities. 
Finally, Figure~\ref{fig:time_to_classification} shows the time to identification as a function of the false-positive rate for the identification and verification settings.

\subsection{Assessing Session Bias}

Using the JuDo data set, we investigate the DeepEyedentification network's ability to  generalize across recording sessions  by comparing its multi-class classification performance on test data taken either from the same sessions that are used for training  or from a new session. 
\begin{figure}[htbp!]
	\subfloat[Evaluation on monocular data. \label{fig:session_bias_mono}]{	
		\includegraphics[width=0.48\textwidth, trim={0.7cm 0 2.2cm 2.4cm}, clip]{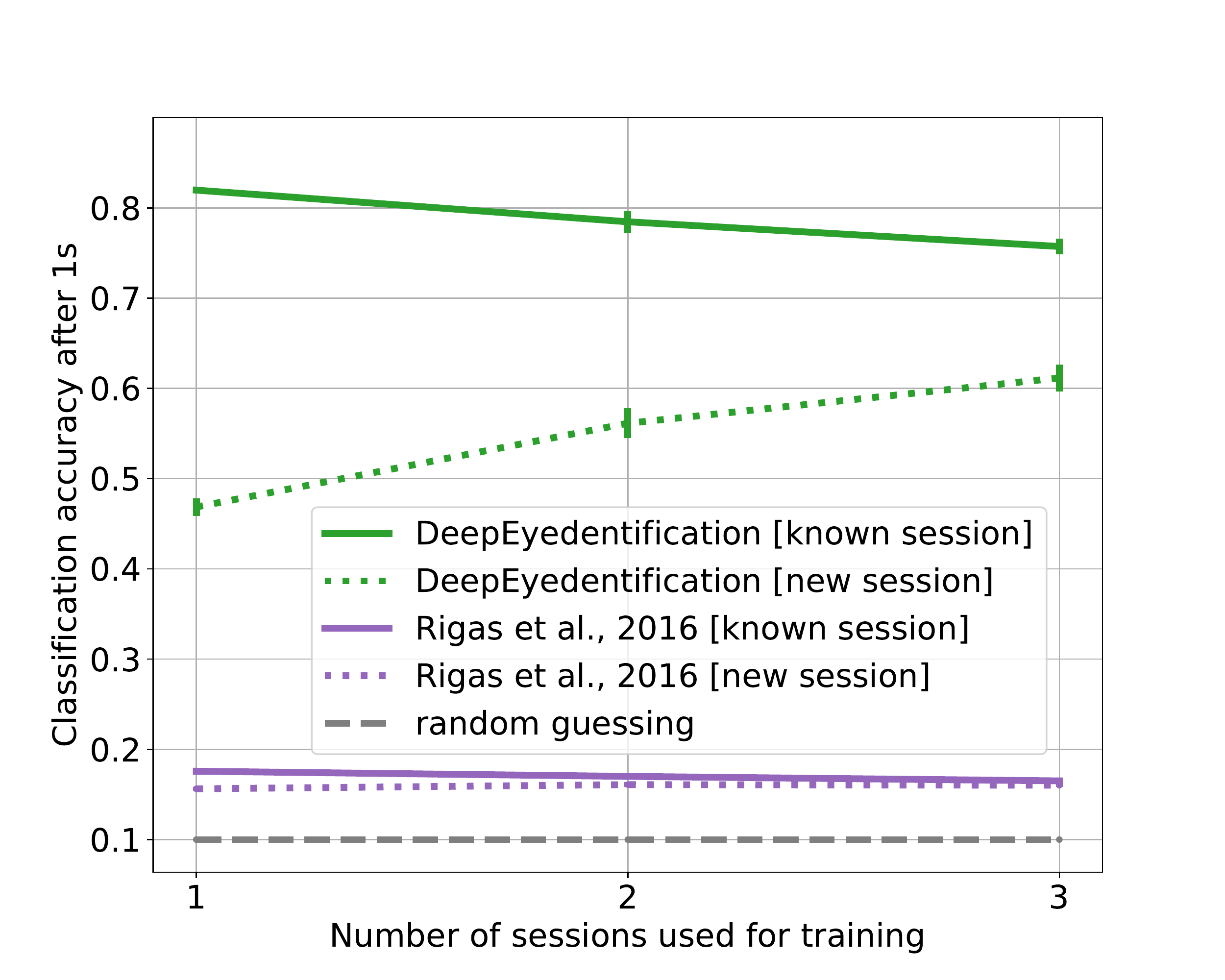}
		}	
		\hfill
	\subfloat[Evaluation on binocular data.	\label{fig:session_bias_binoc}]
	{\includegraphics[width=0.48\textwidth, trim={0.7cm 0 2.2cm 2.4cm}, clip]{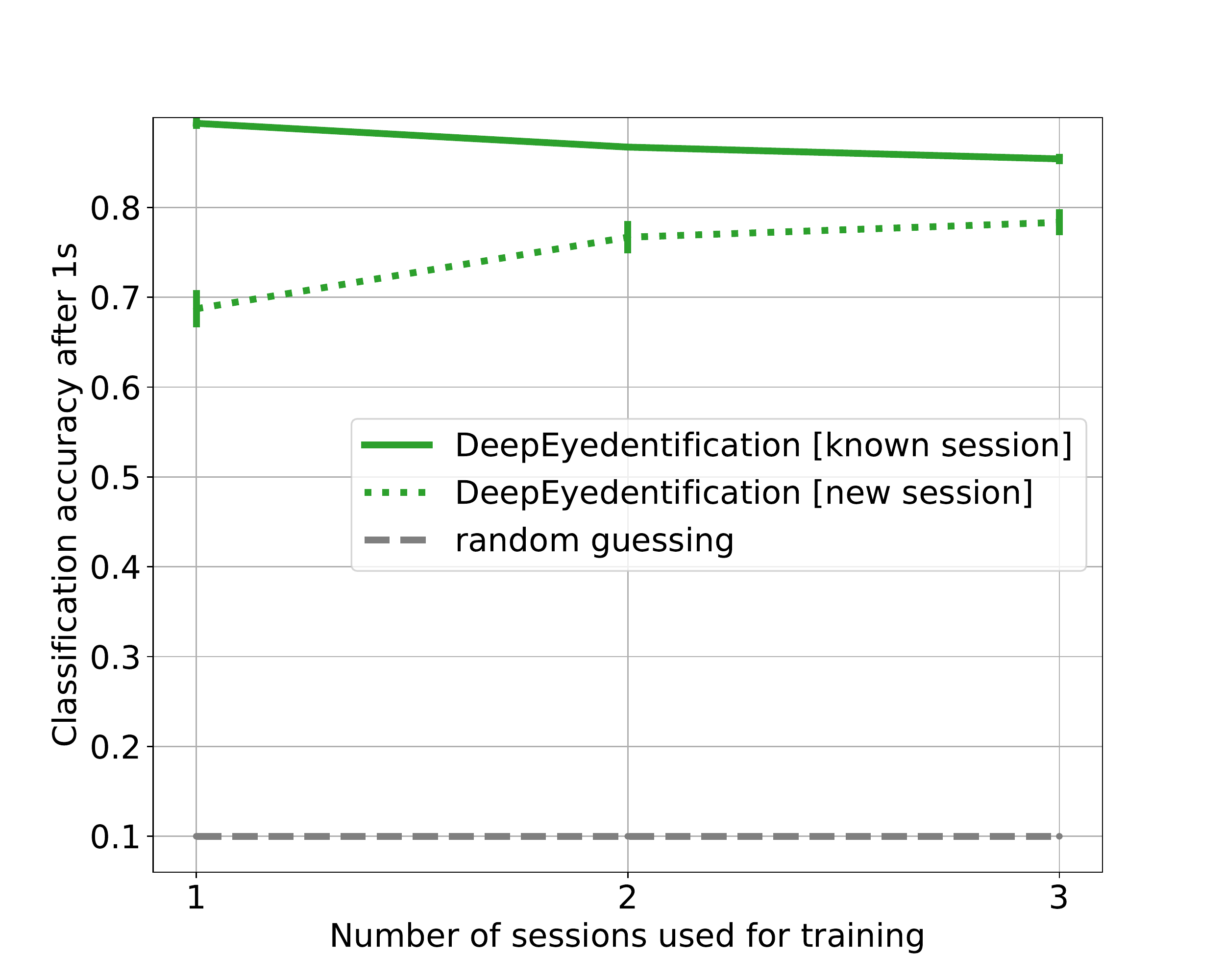}	
	}	
	\caption{Multi-class classification on the JuDo data set.  Categorical accuracy  on one second of test data from either a known or a new  recording session as a function of the number of  sessions used for training with a constant total amount of training data. The results are averaged over ten iterations for each held-out test session.  Error bars show the standard error. 
	}
\end{figure} 
We train the DeepEyedentification network and the reference method that performed best on the Potsdam Textbook Corpus \cite{Rigas2016} on one to three sessions using  the same hyperparameters and learning framework as for the main experiments (see Sections \ref{sec:hyperparams} and \ref{sec:framework}). 
We evaluate the models using leave-one-session-out cross-validation on one held-out session (test on a new session) and on  20\% held-out test data  from the remaining session(s)  (test on a known session). When training on multiple sessions, the amount of training data from each session is reduced such that the total amount of data used for training remains constant. 
Since binocular data is available,  we also evaluate the DeepEyedentification network on binocular data by  applying it independently to synchronous data from both eyes and  averaging the softmax scores of the output layer.  At training, the data from the two eyes are treated as separate instances.

Figure~\ref{fig:session_bias_mono} shows the results for monocular test sequences of one second. 
After one second of input data, the model reaches a classification accuracy of 81.96\% when testing and training it on data from a single session, and an accuracy of up to 61.16\% when training and testing it on  different sessions. 
Increasing the number of training sessions reduces the session bias significantly ($p<0.01$  for one versus  three sessions). 
The model of Rigas {\em et al.}~\cite{Rigas2016} reaches accuracies around 16\% in all settings. 
The use of binocular data (see Figure~\ref{fig:session_bias_binoc}) not only improves the overall performance of the DeepEyedentification network, but also significantly reduces the session bias compared to monocular data ($p<0.01$ for one training session). When being trained on three sessions, the model achieves an accuracy of 78.34\% on a new test session after only one second of input data.

\subsection{Additional Exploratory Experiments}
We briefly summarize the outcome of additional exploratory experiments. First, we explore the behavior of a variant of the DeepEyedentification architecture that has only a single subnet which processes the globally normalized input. This model does not exceed the performance of the fast subnet, which indicates that it extracts only macro-movement patterns. 

Second, we find that adding an input channel that indicates whether a time step is part of a fixation or part of a saccade according to established psychological criteria~\cite{EngbertKliegl2003,EngbertMergenthaler2006} does not improve the model performance. Moreover, forcing the slow subnet to only process movements during fixations and forcing the fast subnet to only process movements during saccades deteriorates the model performance. Our interpretation of this finding is that given the amount of information contained in the training data, an established heuristic categorization of movement types contributes no additional value. 

Lastly, we change the convolutional architecture into a recursive architecture with varying numbers of LSTM units~\cite{Graves2005}. We find that the convolutional architecture consistently outperforms the explored LSTM architectures.

\section{Discussion}\label{sec:discussion}
This section discusses eye movements in relation to other biometric technologies. We discuss relevant qualitative properties of biometric methods: the required level of user interaction, the population for which the method can be applied, attack vectors, and anti-spoofing techniques. 

While fingerprints and hand-vein scans require an explicit user action---placing the finger or the hand on a scanning device---face identification, scanning the iris, and tracking micro-movements of the eye can in principle be performed unobtrusively, without explicit user interaction. Scanning the iris or recording the micro-movements of the eye without requesting the user to step close to a camera would, however, require a camera that offers a sufficiently high resolution over a sufficiently wide field of view. 

Biometric technologies differ with respect to intrinsic limitations of their applicability. For instance, fingerprints are worn down by hard physical labor, iris scanning requires users with small eyes to open their eyes unnaturally wide and is not available for users who wear cosmetic contact lenses. Since micro-movements of the eye are a prerequisite  for vision, this method applies to a large potential user base.

All biometric identification methods can be attacked by acquiring biometric features from an authorized user and replaying the recorded data to the sensor. 
For instance, face identification can be attacked by photographs, video recordings, and 3D masks~\cite{Erdogmus2014}. A replay attack on ocular micro-movement-based identification is theoretically possible but requires a playback device that is able to display a video sequence in the infrared spectrum at a rate of 1,000 frames per second. 
Biometry can similarly be attacked by replaying recorded or artificially generated data during enrollment. For instance, wearing cosmetic contact lenses during enrollment with an iris scanner can cause the scanner to accept other individuals who wear the same contact lens as false positives \cite{Morales2019}. 

Anti-spoofing techniques for all biometric technologies firstly aim at detecting imperfections in replayed data; for example, missing variation in the input over time can indicate a photograph attack. This problem is intrinsically difficult because it is an adversarial problem; an attacker can always minimize artifacts in the replayed data. As an illustration, an attacker can replay a video recording instead of a still image to add liveliness. Liveliness detection is implicitly included in identification based on eye movements. 
Secondly, additional sensors can be added---such as multi-spectral cameras or depth sensors to prevent photograph-based and video-based replay attacks. This of course comes at additional costs and can still be attacked with additional effort, such as by using 3D-printed models instead of photographs. Thirdly, the identification procedure can include a randomized challenge to which the user has to respond. For example, a user can be asked to look at specific positions on a screen~\cite{Maeder2004,Kumar2007,DeLuca2008,Dunphy2008,Weaver2011,Cymek2014}.  
Challenges prevent replay attacks at the cost of obtrusiveness, bypassing them requires a data generator that is able to generate the biometric feature and also respond to the challenge. 
Identification based on movements of the eye is unique: responding to challenges demands neither the user's attention nor a conscious response. Randomized salient stimuli in the field of view immediately trigger an involuntary eye movement that can be validated.

	\section{Conclusion}
	\label{sec:Conclusion}

Our research adds to the list of machine-learning problems for which processing raw input data with a deep CNN greatly improves the  performance over methods that extract engineered features. In this case, the improvement is particularly remarkable and moves a novel biometric-identification technology close to practical applicability. The error rate of the DeepEyedentification network is lower by one order of magnitude and identification is faster by two orders of magnitude compared to the best-performing previously-known method. 

We would like to point out that at this point the embedding layer of DeepEyedentification has been trained with 50 users. 
Nevertheless, it attains a true-positive rate of 60\% at a false-positive rate of 1\% after two seconds of input in the verification setting. 
By comparison, the embedding layer of a current face-identification model that attains a true-positive rate of  95.6\% at a false-positive rate of 1\% has been trained with 9,000 users~\cite{Cao2018}. A recent iris-recognition model attains a true-positive rate of 83.8\% at a false-positive rate of 1\%~\cite{Nalla2017}. 
 This comparison highlights the high potential of  identification based on micro-movements. 

We have developed an approach to processing input that contains signals on vastly different amplitudes. Global normalization squashes the velocities of the most informative, high-frequency but low-amplitude micro-movements to nearly zero, and networks which we train on this type of input do not exceed the performance of the fast subnet. The DeepEyedentification network contains two separately trained subnets that process the same signal scaled such that the velocities of slow movements, in case of the slow subnet, and of fast movements, in case of the fast subnet, populate the input value range.

Biometric identification based on eye movements has many possible fields of application. In contrast to fingerprints  and hand-vein scans, it is unobtrusive. While iris scans fail for cosmetic contact lenses and frequently fail for users with small eyes, it can be applied for all individuals with vision. A replay attack would require a device  able to display 1,000 frames per second in the infrared spectrum. Moreover, replay attacks can be prevented by including a challenge in the form of a visual stimulus in the identification procedure to which the user responds with an involuntary eye movement without assigning attention to the task.

\section*{Acknowledgments}
This work was partially funded by the German Science Foundation under grant  SFB1294, and by the German Federal Ministry of Research and Education  under grant 16DII116-DII. We thank Shravan Vasishth for his support with the data collection.

\bibliographystyle{splncs04}	

\end{document}